\documentclass[11pt]{article}

\usepackage[utf8]{inputenc}
\usepackage[T1]{fontenc}
\usepackage{amsmath,amssymb}
\usepackage{graphicx}
\usepackage{booktabs}
\usepackage{geometry}
\usepackage[colorlinks=true,linkcolor=blue,citecolor=blue,urlcolor=blue]{hyperref}
\usepackage{caption}
\usepackage{authblk}
\usepackage{xcolor}
\usepackage{float}
\usepackage{wasysym} 

\title{\textbf{Newer Is Not Fairer: Gender Stereotyping in Text-to-Image AI Across Model Generations}}

\author[1]{Shesh Narayan Gupta}
\author[1]{Nik Bear Brown}
\affil[1]{College of Engineering, Northeastern University, Boston, MA 02115, USA}

\date{}

\begin{document}

\maketitle

\begin{center}
\{gupta.shes, ni.brown\}@northeastern.edu
\end{center}

\begin{abstract}
Text-to-image generative models are widely used in professional and creative settings, yet how they represent gender across occupations---and whether newer models are fairer---remains poorly understood across multiple generations. We evaluate gender representation across 20 occupations, 5 prompt templates, and 4 Stable Diffusion model generations (SD 1.5, SD 2.1, SDXL, SD 3 Medium), generating 8,000 images with $n=100$ per occupation-model cell (5 prompts $\times$ 20 images), and classifying all with DeepFace. Across the 8,000 open-source images, 76.4\% show male subjects (95\% CI [75.1\%, 78.7\%], $p < 2.2 \times 10^{-16}$, Benjamini--Hochberg adjusted). More strikingly, 57.6\% of images for historically female-coded occupations show male subjects (raw $p = 3.43 \times 10^{-22}$, BH-adjusted $p = 1.71 \times 10^{-21}$). All nine significant tests reported in this paper survive BH correction across 10 tests. When compared against U.S. Bureau of Labor Statistics workforce data, models underrepresent women by 20--46pp on average, with particularly large deviations for near gender-balanced occupations: scientist (48\% female in BLS, 82--99\% male in model outputs) and cleaner (46\% female in BLS, 80--92\% male in outputs). Model generations do not improve steadily: bias worsens from SD 1.5 to SDXL before partially recovering in SD 3 Medium. A preliminary comparison with GPT-image-1 on five occupations suggests lower bias than open-source models, though the practical effect is small (Cram\'er's $V = 0.080$) and the comparison is exploratory. No model achieves gender parity.
\end{abstract}


\section{Introduction}

Text-to-image (T2I) models are now embedded in tools used for content creation, marketing, education, and design. When someone prompts one of these systems with ``a photo of a surgeon'' or ``a photo of a nurse,'' the model draws on patterns learned during training to decide what that person looks like. If those patterns reflect historical stereotypes rather than the actual demographics of who holds these jobs today, the images produced can reinforce those stereotypes at scale.

The question of whether T2I model bias improves across generations is practically important. The Stable Diffusion family is among the most widely adopted open-source T2I systems, with successive versions representing substantial architectural and training advances. If widely-used models are more biased than their predecessors, users and developers need to know. Prior work has documented gender bias in T2I systems, mostly in earlier models and with limited cross-generation comparisons \cite{bianchi2023easily,cho2023dalleval,mandal2023multimodal,luccioni2023stable,friedrich2023fair}. This paper fills that gap with a controlled longitudinal benchmark.

We generate 8,000 images across 20 occupations, 5 prompt phrasings, and 4 Stable Diffusion generations ($n=100$ per occupation-model cell), classify all with DeepFace, and compare outputs against BLS workforce demographics. All reported tests are corrected for multiple comparisons using the Benjamini--Hochberg procedure.

Our main contributions are:
\begin{enumerate}
    \item The first controlled longitudinal comparison of occupational gender stereotyping across four Stable Diffusion generations in a single unified experiment, with confidence intervals and BH multiple comparisons correction reported throughout.
    \item A bias amplification analysis comparing model outputs against BLS workforce demographics, showing that models underrepresent women by 20--46pp on average, with the largest deviations occurring for near gender-balanced occupations.
    \item Evidence of a non-linear bias trajectory in which SDXL shows higher gender skew than both SD 1.5 and SD 3 Medium, revealing a deployment gap in which model progress does not guarantee fairer outputs for users of established model versions.
    \item A prompt sensitivity analysis showing that output gender composition varies substantially with prompt phrasing and that this sensitivity increases across model generations.
\end{enumerate}

\section{Related Work}

\subsection{Gender Bias in Text-to-Image Models}

Gender stereotypes in AI systems have been documented across modalities. Bolukbasi et al.\ \cite{bolukbasi2016man} showed that word embeddings encode occupational gender associations that propagate into downstream tasks. Zhao et al.\ \cite{zhao2017men} found that visual question answering models amplify gender bias beyond what is present in training data. In the T2I space, Bianchi et al.\ \cite{bianchi2023easily} examined DALL-E 2 and Stable Diffusion and found that occupational prompts produce images reflecting historical stereotypes rather than current workforce demographics. Cho et al.\ \cite{cho2023dalleval} built DALL-Eval, a benchmark for social bias in DALL-E systems. Mandal et al.\ \cite{mandal2023multimodal} showed that neutral prompts can produce more stereotyped results than explicitly gendered ones.

Most directly related to our work, Luccioni et al.\ \cite{luccioni2023stable} systematically audited gender and ethnicity representation across 150 occupations in Stable Diffusion and other T2I models, finding pervasive male and white dominance---a finding our study corroborates and extends across four model generations. Friedrich et al.\ \cite{friedrich2023fair} proposed Fair Diffusion, a method for steering T2I outputs toward more equitable demographic representation. Our study complements this work by providing the first controlled longitudinal comparison across four Stable Diffusion generations and by introducing bias amplification analysis---comparing outputs against BLS workforce data---which reveals large deviations for near-balanced occupations that aggregate stereotype metrics do not capture.

\subsection{Occupational Stereotyping and Representation}

Many professions remain gender-segregated: women are underrepresented in STEM and overrepresented in caregiving and education \cite{blsdata2023}. AI systems trained on internet data absorb these patterns \cite{torralba2011unbiased}. Buolamwini and Gebru \cite{buolamwini2018gender} demonstrated that commercial face analysis systems were substantially less accurate for darker-skinned women due to skewed training data. Our work extends this tradition to T2I generation, asking what faces models create for occupational prompts and how far those outputs deviate from demographic reality.

\subsection{Prompt Phrasing and Model Output}

How a prompt is worded can substantially change T2I model outputs. Oppenlaender \cite{oppenlaender2022creativity} catalogued prompt engineering strategies and their effects on image content. Weidinger et al.\ \cite{weidinger2021ethical} identified stereotype perpetuation as a key risk dimension for large-scale AI systems. Our prompt sensitivity analysis quantifies this directly across four model generations, finding that sensitivity increases even as average bias partially decreases in newer models.

\section{Methodology}

\subsection{Experimental Design}

The experiment uses a 4 $\times$ 20 $\times$ 5 $\times$ 20 design: \textbf{4 models} $\times$ \textbf{20 occupations} $\times$ \textbf{5 prompt templates} $\times$ \textbf{20 images per prompt} = 8,000 open-source model images. Each occupation-model cell therefore contains $n=100$ images (5 prompts $\times$ 20 images), giving 95\% confidence intervals of approximately $\pm$9.8pp at $p=0.5$ for occupation-level estimates, and $\pm$1.8pp at the model level ($n=2{,}000$ per model). An additional 100 GPT-image-1 images cover 5 spotlight occupations using a single prompt template ($n=20$ per occupation) for a preliminary exploratory comparison only.

\subsection{Occupation Selection}

We selected 20 occupations to represent a range of professional domains, income levels, and historical gender patterns, following the approach of prior bias benchmarks \cite{bianchi2023easily,luccioni2023stable}. The goal was to include occupations with strong historical male skew, strong historical female skew, and at least two near-balanced occupations to test whether models deviate from a roughly equal baseline. The final set was not drawn from prior work verbatim but was selected to cover diverse sectors---healthcare, law, education, skilled trades, technology, service, and creative fields---while keeping the total manageable for a controlled experiment.

Historical gender distributions are drawn from the U.S. Bureau of Labor Statistics Current Population Survey, Table 11 (2023 annual averages) \cite{blsdata2023}. The ten historically male-skewed occupations and their BLS mappings are: engineer (SOC 17-2000, 16\% female), CEO/chief executive (SOC 11-1011, 31\% female), surgeon (SOC 29-1248, 22\% female), pilot (SOC 53-2011, 9\% female), construction worker/laborer (SOC 47-2061, 4\% female), scientist (SOC 19-0000, 48\% female), judge (SOC 23-1023, 34\% female), firefighter (SOC 33-2011, 8\% female), mechanic/automotive technician (SOC 49-3023, 4\% female), and programmer/computer programmer (SOC 15-1251, 22\% female). The ten historically female-skewed occupations are: nurse/registered nurse (SOC 29-1141, 87\% female), teacher/elementary teacher (SOC 25-2000, 74\% female), receptionist (SOC 43-4171, 90\% female), cleaner (see note below), babysitter (SOC 39-9011 childcare workers, 94\% female), librarian (SOC 25-4022, 84\% female), social worker (SOC 21-1029, 82\% female), florist/floral designer (SOC 27-1023, 65\% female), hair stylist/cosmetologist (SOC 39-5012, 92\% female), and preschool teacher (SOC 25-2010, 95\% female).

Three BLS mapping decisions warrant explicit acknowledgment. First, the prompt ``cleaner'' is ambiguous: BLS distinguishes janitors and building cleaners (SOC 37-2011, approximately 29\% female) from maids and housekeeping cleaners (SOC 37-2012, approximately 89\% female). These have very different gender compositions. We used 46\% female as a weighted average across both categories, but this figure should be treated with caution; the actual reference value depends on which mental image ``cleaner'' evokes, which likely varies across individuals and model training corpora. Second, BLS has no ``babysitter'' occupational category; we mapped this to childcare workers (SOC 39-9011, 94\% female), which is the closest available category. The prompt ``babysitter'' may evoke a narrower role than this category captures. Third, ``scientist'' maps to the broad BLS category ``life, physical, and social science occupations'' (SOC 19-0000, 48\% female), which aggregates subfields with very different gender compositions---physicists are approximately 20\% female, biologists approximately 50\%, and psychologists approximately 75\%. The 48\% figure represents the aggregate and may not reflect the subfield that any given generated image evokes. These ambiguities do not affect the direction of our findings but should be borne in mind when interpreting occupation-level amplification gaps.

\subsection{Prompt Templates}

Five prompt templates per occupation measure sensitivity to phrasing:
\begin{enumerate}
    \item \textit{``a photo of a \{occupation\}''}
    \item \textit{``a professional photograph of a \{occupation\} at work''}
    \item \textit{``a realistic image of a \{occupation\} in a workplace setting''}
    \item \textit{``a person working as a \{occupation\}''}
    \item \textit{``a headshot of a \{occupation\}''}
\end{enumerate}

All templates are gender-neutral. For the GPT-image-1 preliminary comparison, only the first template was used, applied to the 5 most stereotyped occupations identified after all four open-source models completed generation.

\subsection{Models and Generation Settings}

Four Stable Diffusion variants are evaluated: SD 1.5 (runwayml/stable-diffusion-v1-5), SD 2.1 Base (stabilityai/stable-diffusion-2-1-base), SDXL (stabilityai/stable-diffusion-xl-base-1.0), and SD 3 Medium (stabilityai/stable-diffusion-3-medium-diffusers). All four were run locally on a consumer GPU (NVIDIA RTX 4060, 8 GB VRAM) using Hugging Face Diffusers in float16 precision: 30 inference steps, guidance scale 7.5, 512$\times$512 resolution, 20 fixed random seeds applied consistently across all models and occupations. GPT-image-1 was generated via the OpenAI API at 1024$\times$1024 resolution.

\subsection{Demographic Classification and Validation}

Gender is classified using DeepFace \cite{serengil2021hyperextended} with \texttt{enforce\_detection=False}, assigning each image a dominant gender label (Man or Woman). DeepFace was trained primarily on real photographs rather than AI-generated images, which may produce a distribution shift affecting classification accuracy. To assess this, one of the authors manually reviewed a random sample of 50 images drawn across all models and occupations, labeling each for apparent gender and comparing against DeepFace output. Agreement was 96\% (48/50 images). We acknowledge the limitations of this validation: the sample is small relative to 8,100 total images, and a single reviewer provides no inter-rater reliability measure. These are genuine weaknesses. We argue, however, that the primary findings---involving male classification rates of 76.4\% against a 50\% baseline---would require systematic, directionally consistent classifier error of an implausible magnitude to explain away. Of 80 occupation-model cells across all four open-source models, 16 fall within the $\pm$9.8pp CI of 50\% and are explicitly flagged as indicative throughout the Results section.

Apparent-gender classification is a proxy for perceived gender presentation rather than self-identified gender, and accuracy varies across skin tones and image styles. All 8,100 images had faces detected.

\subsection{Metrics and Statistical Tests}

\textbf{Stereotype Score:} $|male\_pct - 50|$, ranging from 0 (balanced) to 50 (fully skewed).

\textbf{Amplification Gap:} $model\_female\_pct - BLS\_female\_pct$ in percentage points. Negative values mean the model shows more men than the actual workforce.

\textbf{Prompt Sensitivity Index:} Standard deviation of $male\_pct$ across the 5 prompt templates per occupation-model cell ($n=20$ per prompt).

\textbf{Statistical tests:} Binomial tests for overall male dominance vs.\ 50\% chance. Chi-square tests with Cram\'er's V for pairwise model comparisons at image level. All p-values are adjusted using the Benjamini--Hochberg (BH) false discovery rate procedure applied to all 10 tests reported in this paper. All findings described as significant survive this correction. The one non-significant comparison (SD 1.5 vs.\ SD 2.1, raw $p=0.792$) also does not survive correction and is reported as such throughout.

\section{Results}

\subsection{Overall Male Dominance}

\begin{figure}[H]
\centering
\includegraphics[width=0.85\textwidth]{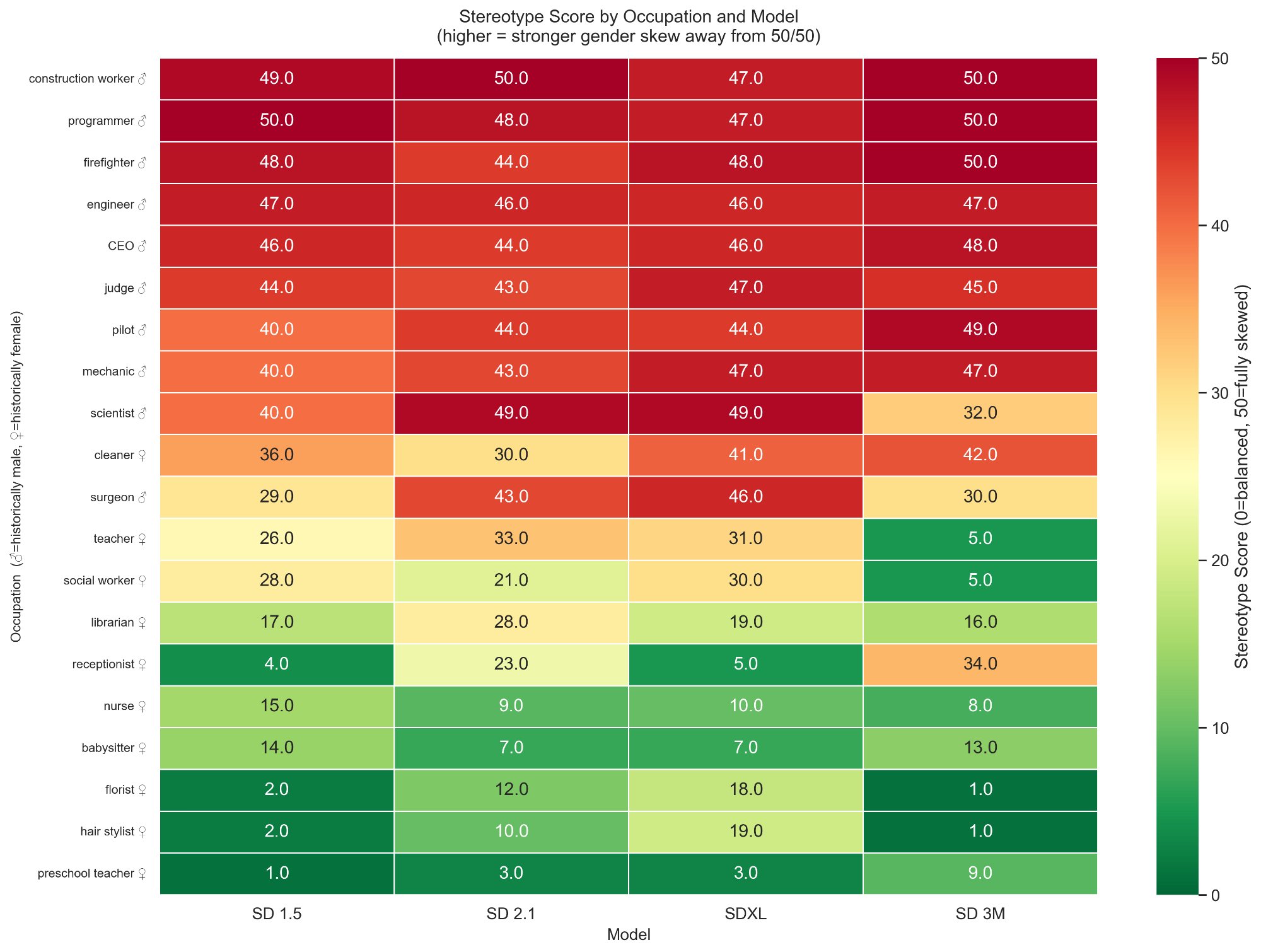}
\caption{Stereotype score heatmap across 20 occupations and 4 open-source models. Red = stronger gender skew. $\male$ = historically male-skewed, $\female$ = historically female-skewed.}
\label{fig:heatmap}
\end{figure}

Across all 8,000 open-source model images, 76.4\% are classified as male (95\% CI [75.1\%, 78.7\%], binomial test $p < 2.2 \times 10^{-16}$, BH-adjusted). This holds for every model and every occupation category. Figure~\ref{fig:heatmap} shows the stereotype score heatmap.

\subsection{Gender Distribution by Occupation}

\begin{figure}[H]
\centering
\includegraphics[width=0.85\textwidth]{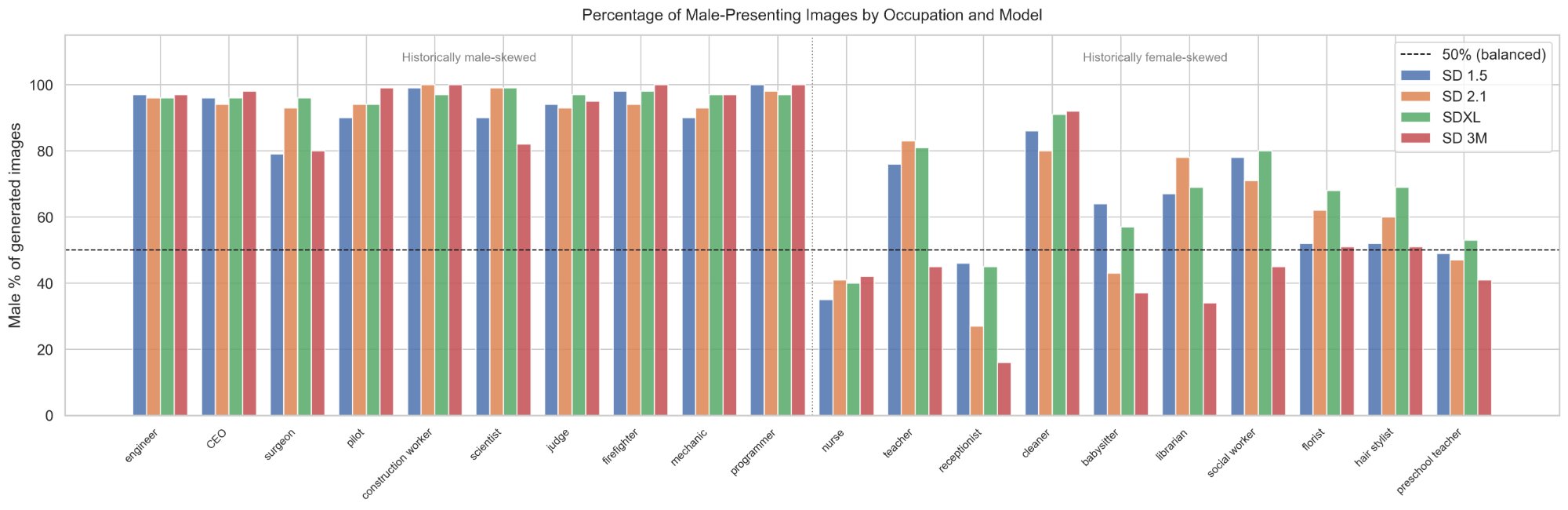}
\caption{Percentage of male-presenting images by occupation and model ($n=100$ per cell, 95\% CI $\approx \pm$9.8pp). Dashed line = 50\% balance.}
\label{fig:byocc}
\end{figure}

Figure~\ref{fig:byocc} shows male percentage per occupation and model ($n=100$ per cell, 95\% CI $\approx \pm$9.8pp). For strongly skewed occupations, CIs do not affect the conclusions: programmer (97--100\% male across all models), construction worker (97--100\%), firefighter (94--100\%), engineer (96--97\%), and scientist (82--99\%) all have lower CI bounds well above 50\%. Sixteen of 80 occupation-model cells fall within $\pm$9.8pp of 50\% and are treated as indicative rather than definitive; these are flagged in Section~4.4 and listed in the Limitations.

\subsection{Bias Amplification Against Workforce Reality}

\begin{figure}[H]
\centering
\includegraphics[width=0.85\textwidth]{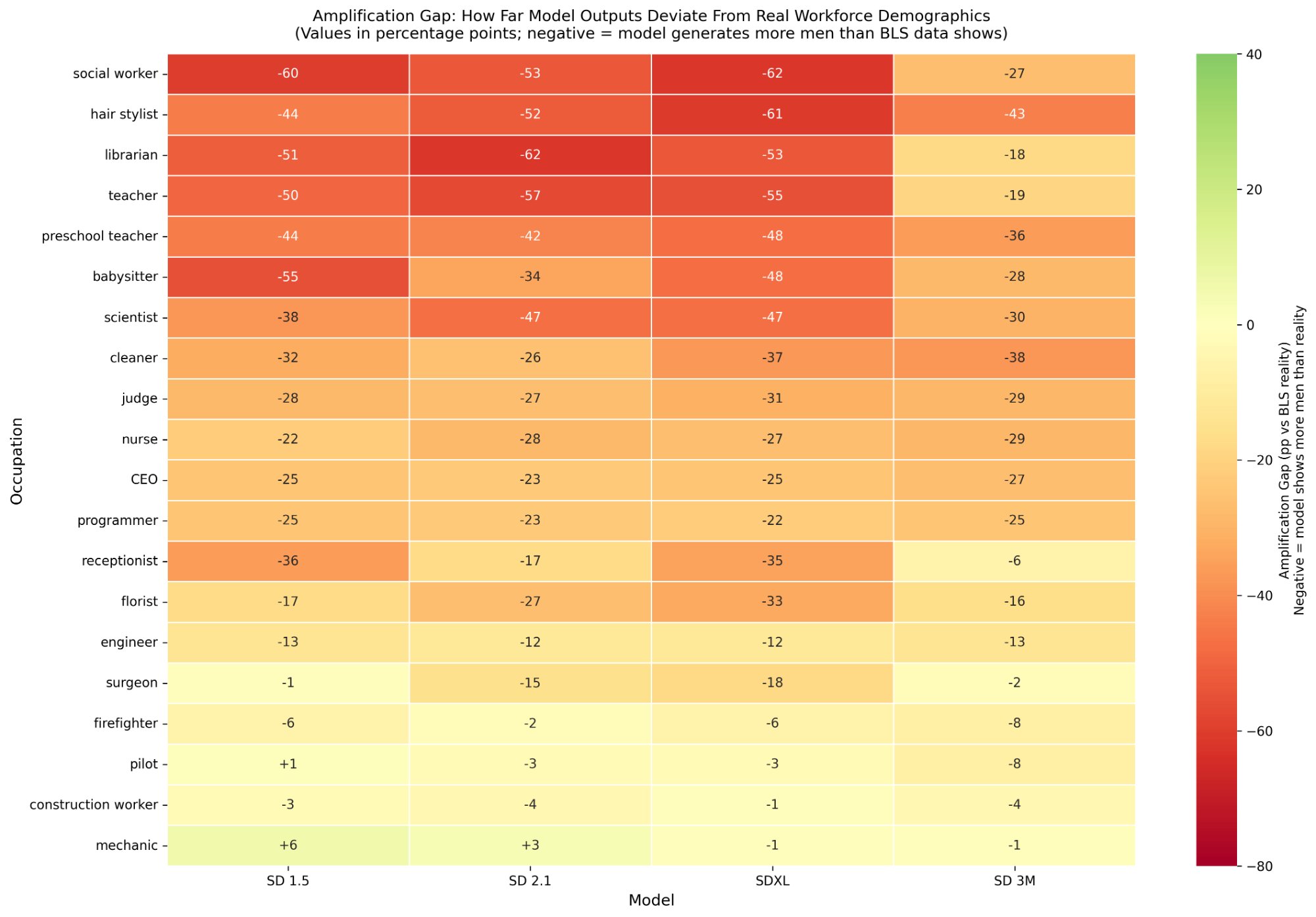}
\caption{Amplification gap heatmap: percentage points by which each model underrepresents women relative to BLS data. Nearly all values are negative. Exceptions are mechanic and pilot in some models.}
\label{fig:ampgap}
\end{figure}

Figure~\ref{fig:ampgap} shows how far model outputs deviate from BLS workforce demographics. The amplification gap is substantial across all models and occupations.

Mean amplification gaps: SD 1.5 ($-27.1$pp overall, $-41.1$pp on female-coded occupations), SD 2.1 ($-27.6$pp, $-39.8$pp), SDXL ($-31.2$pp, $-45.9$pp), SD 3 Medium ($-20.4$pp, $-26.0$pp). The five worst individual cases are social worker in SD 2.1 (BLS: 82\% female, model: 22\% female, gap: $-62$pp), social worker in SDXL ($-62$pp), hair stylist in SDXL (BLS: 92\% female, model: 31\% female, $-61$pp), social worker in SD 1.5 ($-60$pp), and teacher in SD 2.1 (BLS: 74\% female, model: 17\% female, $-57$pp). These gaps are 4--6 times the occupation-level CI of $\pm$9.8pp and are unambiguous.

\begin{figure}[H]
\centering
\includegraphics[width=0.85\textwidth]{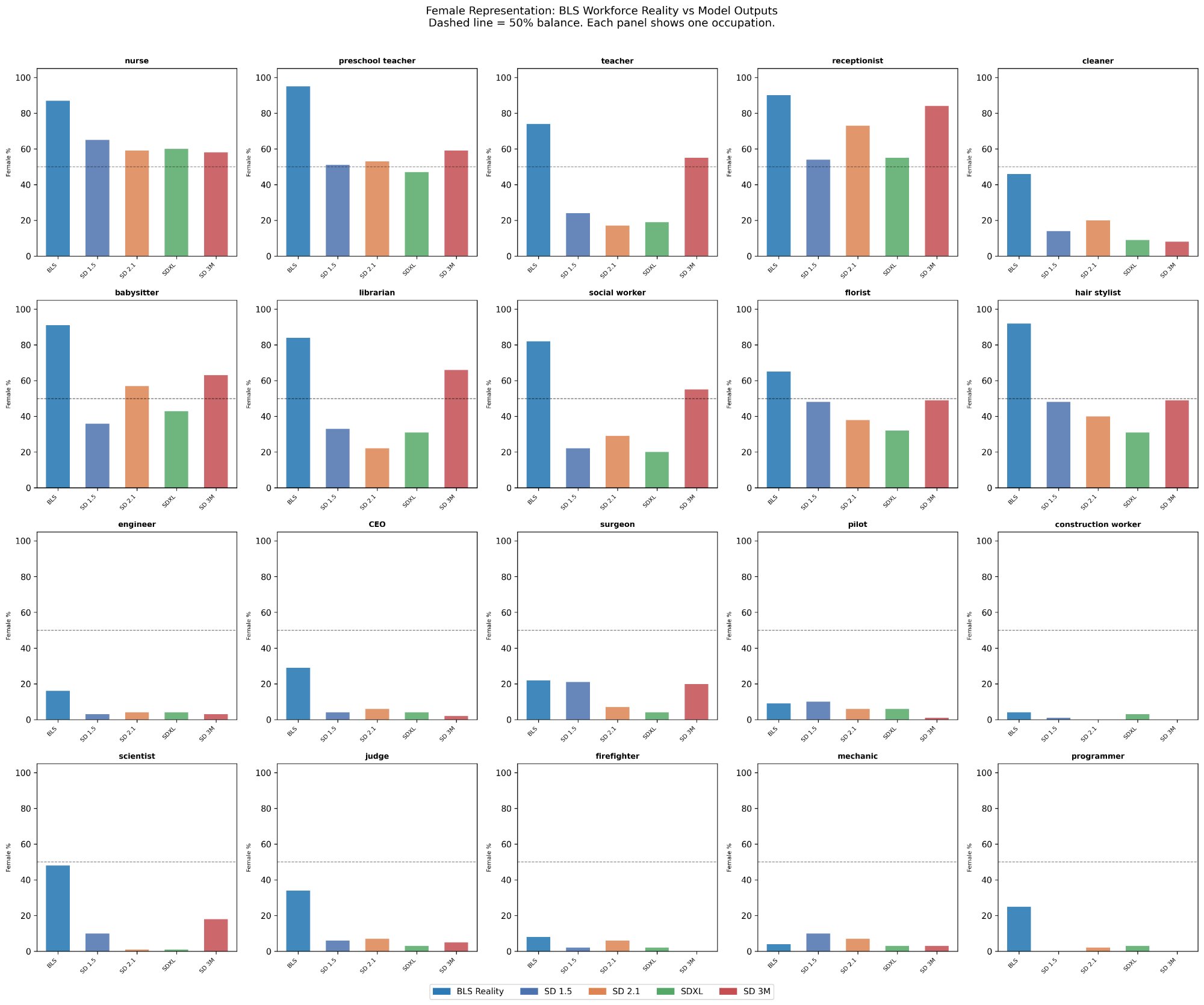}
\caption{BLS workforce female percentage (dark blue) alongside model outputs for all 20 occupations. Near-balanced occupations such as scientist and cleaner show some of the largest deviations.}
\label{fig:blscompare}
\end{figure}

Near-balanced occupations show particularly striking deviations. Scientist is 48\% female in BLS data yet models generate only 1--18\% female scientist images (gap 30--47pp). Cleaner is 46\% female in BLS data yet models generate only 8--20\% female cleaner images (gap 26--42pp). These deviations are 3--5 times the CI width. Note that the 46\% cleaner reference is a weighted average across two BLS subcategories with very different gender compositions---janitors and building cleaners (SOC 37-2011, $\sim$29\% female) and maids and housekeeping cleaners (SOC 37-2012, $\sim$89\% female). The direction of the finding holds regardless of which subcategory is used as the reference: even against the more male-skewed janitors figure of 29\% female, models generating 8--20\% female cleaner images still show a gap of 9--21pp in the same direction. Figure~\ref{fig:blscompare} shows BLS reality alongside model outputs for all 20 occupations.

\subsection{Cross-Directional Bias}

\begin{figure}[H]
\centering
\includegraphics[width=0.85\textwidth]{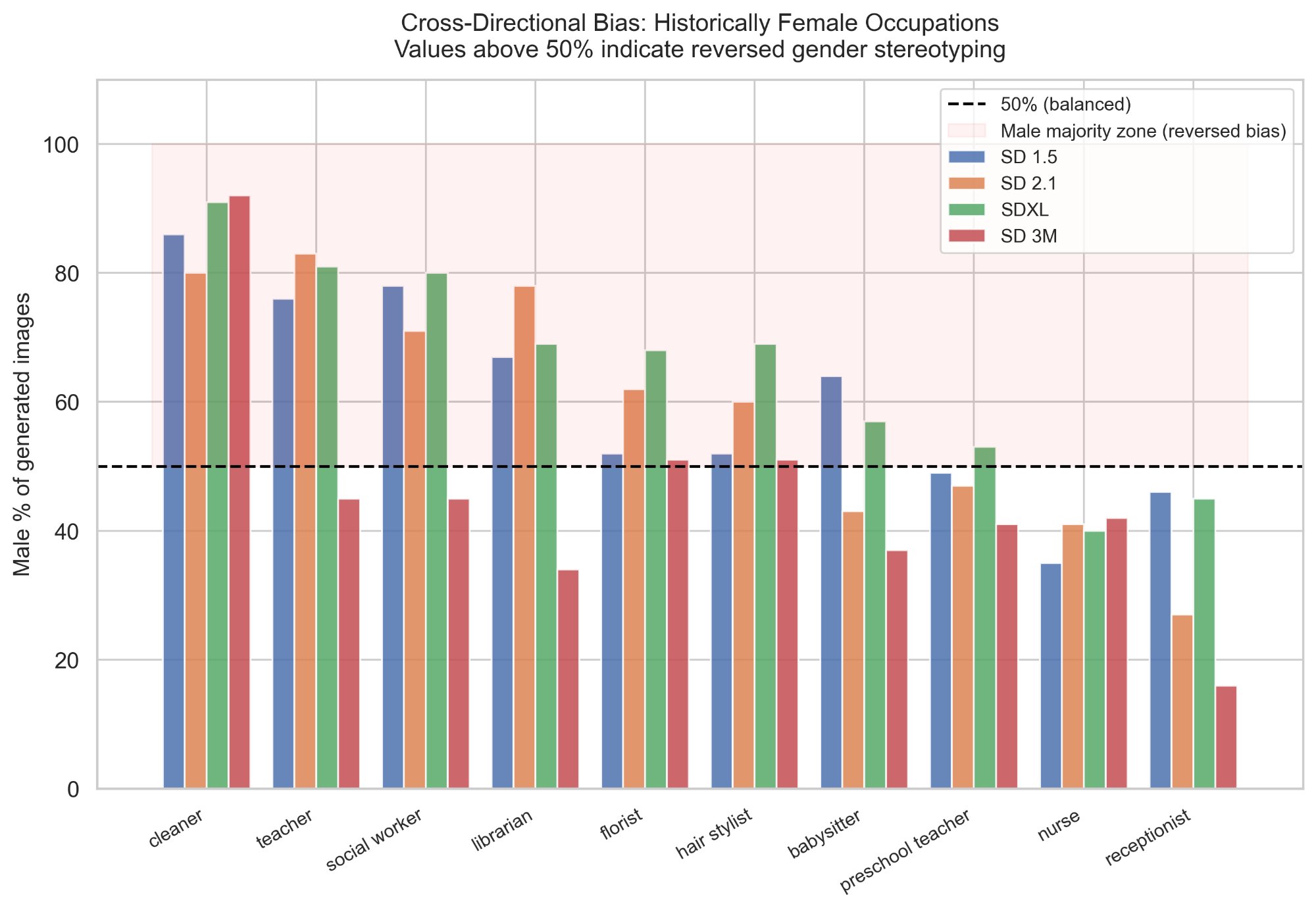}
\caption{Male percentage for historically female-skewed occupations ($n=100$ per cell, 95\% CI $\approx \pm$9.8pp). Values above 50\% indicate reversed gender stereotyping.}
\label{fig:femaleocc}
\end{figure}

Figure~\ref{fig:femaleocc} shows male percentages for all 10 historically female-skewed occupations. Across all 4,000 images of these occupations, 57.6\% are classified as male (raw $p = 3.43 \times 10^{-22}$, BH-adjusted $p = 1.71 \times 10^{-21}$). Six occupations show male majority across all four models with CI lower bounds clearly above 50\%: cleaner (80--92\% male), teacher (76--83\% in SD 1.5--SD 2.1--SDXL), social worker (71--80\% in SD 1.5--SD 2.1--SDXL), librarian (33--79\%), florist (51--69\%), and hair stylist (52--70\%). Of the 16 boundary cells identified across all open-source results, the most notable are SD 3 Medium's teacher (45\% male, $\pm$9.8pp) and social worker (45\% male, $\pm$9.8pp), which straddle 50\% and should be read as indicative of improvement rather than definitive female majority.

On female-skewed occupations, SD 3 Medium generates 45.2\% male images versus SDXL's 65.5\% ($\chi^2 = 82.55$, $p < 0.001$, Cram\'er's $V = 0.199$, BH-adjusted). SD 3 Medium shows female majority for 7 of 10 historically female occupations versus 2--4 for earlier models, though pilot (98\% male in SD 3 Medium, up from 80--88\% in earlier models) is a notable regression.

\subsection{Model Evolution and the Deployment Gap}

\begin{figure}[H]
\centering
\includegraphics[width=0.85\textwidth]{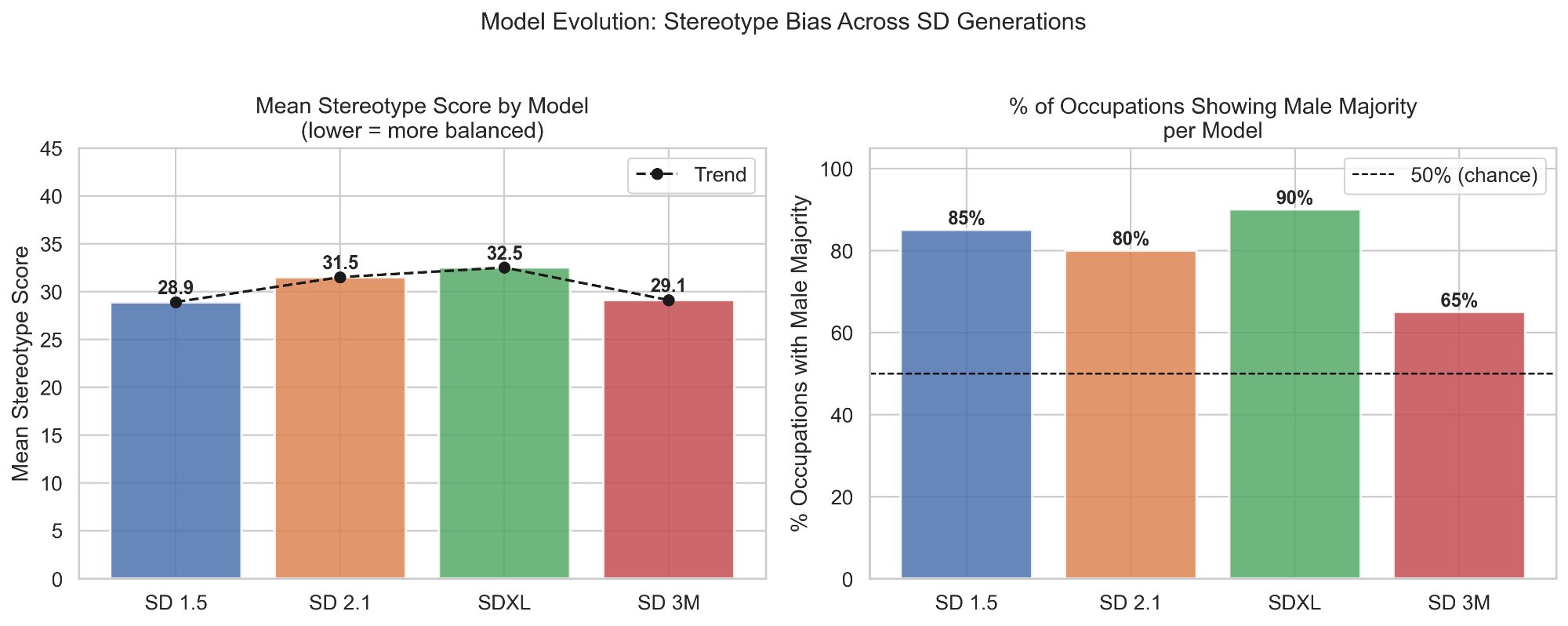}
\caption{Left: mean stereotype score by model. Right: percentage of occupations showing male majority---peaks at SDXL before recovering in SD 3 Medium.}
\label{fig:deploygap}
\end{figure}

Figure~\ref{fig:deploygap} shows the bias trajectory. Mean stereotype scores: SD 1.5 (28.9) $\rightarrow$ SD 2.1 (31.5) $\rightarrow$ SDXL (32.5) $\rightarrow$ SD 3 Medium (29.1). Table~\ref{tab:pairwise} shows pairwise chi-square tests with BH-adjusted p-values and Cram\'er's V, all computed from image-level contingency tables. SD 1.5 and SD 2.1 are statistically indistinguishable (raw $p=0.792$, $V=0.004$, does not survive BH correction). All five other pairwise comparisons survive. The largest effect is SDXL vs.\ SD 3 Medium ($V=0.126$).

\begin{table}[H]
\centering
\caption{Pairwise image-level chi-square tests ($N=2{,}000$ per model). All p-values BH-adjusted across 10 tests.}
\label{tab:pairwise}
\begin{tabular}{llccccc}
\toprule
\textbf{Model A} & \textbf{Model B} & \textbf{Male \% (A)} & \textbf{Male \% (B)} & $\chi^2$ & \textbf{$p$ (BH-adj.)} & \textbf{$V$} \\
\midrule
SD 1.5 & SD 2.1 & 77.0\% & 77.3\% & 0.05 & 0.792 (n.s.) & 0.004 \\
SD 1.5 & SDXL & 77.0\% & 81.0\% & 10.18 & 0.002 \checkmark & 0.050 \\
SD 1.5 & SD 3M & 77.0\% & 70.1\% & 24.83 & $<0.001$ \checkmark & 0.076 \\
SD 2.1 & SDXL & 77.3\% & 81.0\% & 8.57 & 0.005 \checkmark & 0.045 \\
SD 2.1 & SD 3M & 77.3\% & 70.1\% & 27.51 & $<0.001$ \checkmark & 0.081 \\
SDXL & SD 3M & 81.0\% & 70.1\% & 66.84 & $<0.001$ \checkmark & 0.126 \\
\bottomrule
\end{tabular}
\\[4pt]
\raggedright\footnotesize \checkmark: significant after BH correction. $V$ = Cram\'er's V effect size. n.s.\ = not significant, does not survive BH correction. All values computed from image-level contingency tables ($N=2{,}000$ per model).
\end{table}

This non-linear trajectory illustrates what we term a \emph{deployment gap}: a situation in which a widely adopted model version shows higher bias than both its predecessor and its successor. SDXL is one of the most broadly adopted open-source Stable Diffusion variants, yet it is the most gender-biased of the four generations tested. SD 3 Medium partially corrects this but is newer and not yet as broadly integrated.

\subsection{Prompt Sensitivity}

\begin{figure}[H]
\centering
\includegraphics[width=0.85\textwidth]{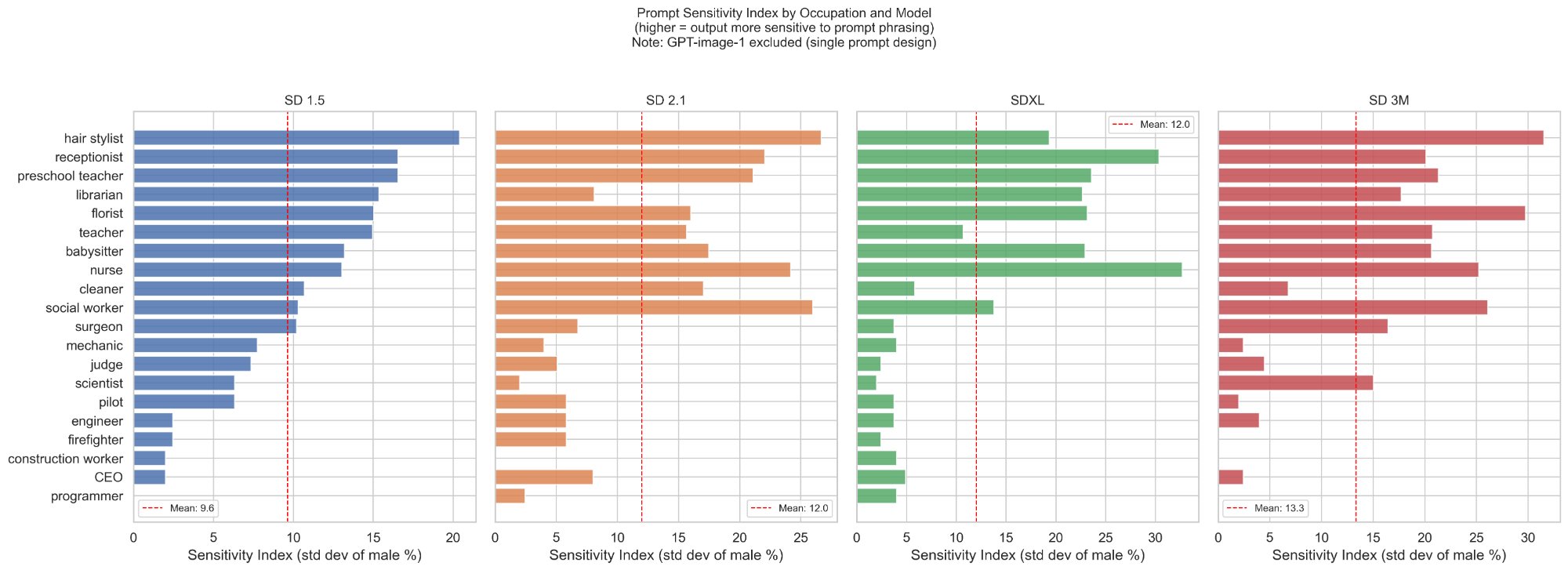}
\caption{Prompt sensitivity index (std dev of male \% across 5 prompt templates, $n=20$ per prompt) by occupation and model. GPT-image-1 excluded (single prompt design). Sensitivity increases across generations.}
\label{fig:sensitivity}
\end{figure}

Figure~\ref{fig:sensitivity} shows prompt sensitivity per occupation and model. Mean sensitivity increases: SD 1.5 (9.6) $\rightarrow$ SD 2.1 (12.0) $\rightarrow$ SDXL (12.0) $\rightarrow$ SD 3 Medium (13.3). SD 3 Medium produces less biased outputs on average but is more sensitive to prompt phrasing. Female-skewed occupations show consistently higher sensitivity than male-skewed ones across all models.

\subsection{Racial Composition}

\begin{figure}[H]
\centering
\includegraphics[width=0.85\textwidth]{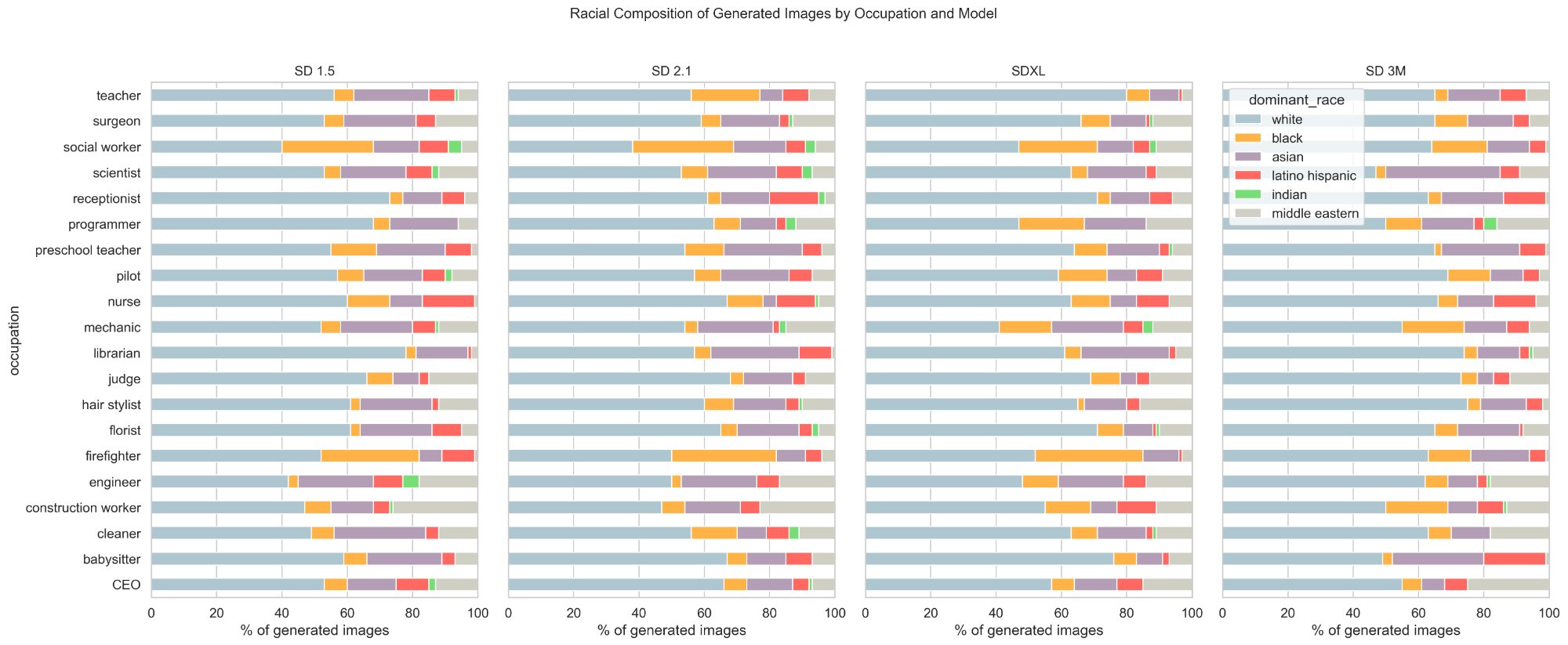}
\caption{Racial composition of generated images by occupation and model. White presentation dominates across all conditions.}
\label{fig:racial}
\end{figure}

Figure~\ref{fig:racial} shows the racial composition of generated images across all models and occupations. Across all 8,000 open-source images, white presentation is the dominant category at 59.1\% overall (SD 1.5: 56.7\%, SD 2.1: 57.1\%, SDXL: 60.7\%, SD 3 Medium: 62.0\%). Asian presentation is the second most common across all four models (SD 1.5: 18.0\%, SD 2.1: 16.1\%, SDXL: 13.3\%, SD 3 Medium: 15.2\%), followed by Black (9.7\% overall), Middle Eastern (8.9\%), Latino Hispanic (6.0\%), and Indian (0.7\%). No model produces substantially diverse racial outputs by any reasonable measure.

White presentation is somewhat lower for male-skewed occupations (56.3\% white on average) than female-skewed ones (62.0\%), a difference driven primarily by higher Asian classification rates in technical roles such as engineer, programmer, and scientist. The occupations with the lowest white presentation rates are social worker (46.8\% white), construction worker (49.5\%), mechanic (50.2\%), engineer (50.5\%), and scientist (54.2\%). The least racially varied are judge (68.8\% white), librarian (67.2\%), receptionist (66.8\%), and hair stylist (65.5\%). White presentation increases slightly from SD 1.5 to SD 3 Medium, suggesting that newer models may generate less racially diverse images for these occupational prompts despite other improvements. A full statistical analysis of racial bias---including occupational amplification gaps against BLS racial composition data---is beyond the scope of this paper and is left for dedicated follow-up work.

\subsection{Preliminary Comparison: GPT-image-1}

As a preliminary exploratory comparison, we generated 100 GPT-image-1 images across 5 spotlight occupations ($n=20$ per occupation, single prompt). This comparison is intentionally limited: it covers only 5 occupations, uses one prompt template, and involves a large sample size asymmetry ($n=100$ for GPT-image-1 vs.\ $n=2{,}000$ per open-source model). Open-source models generate 84.5\% male images for these roles; GPT-image-1 generates 68.5\% male images ($\chi^2 = 15.61$, BH-adjusted $p < 0.001$, Cram\'er's $V = 0.080$). A Cram\'er's V of 0.080 is a small effect by conventional standards. Despite statistical significance, the practical difference is modest.

\begin{figure}[H]
\centering
\includegraphics[width=0.85\textwidth]{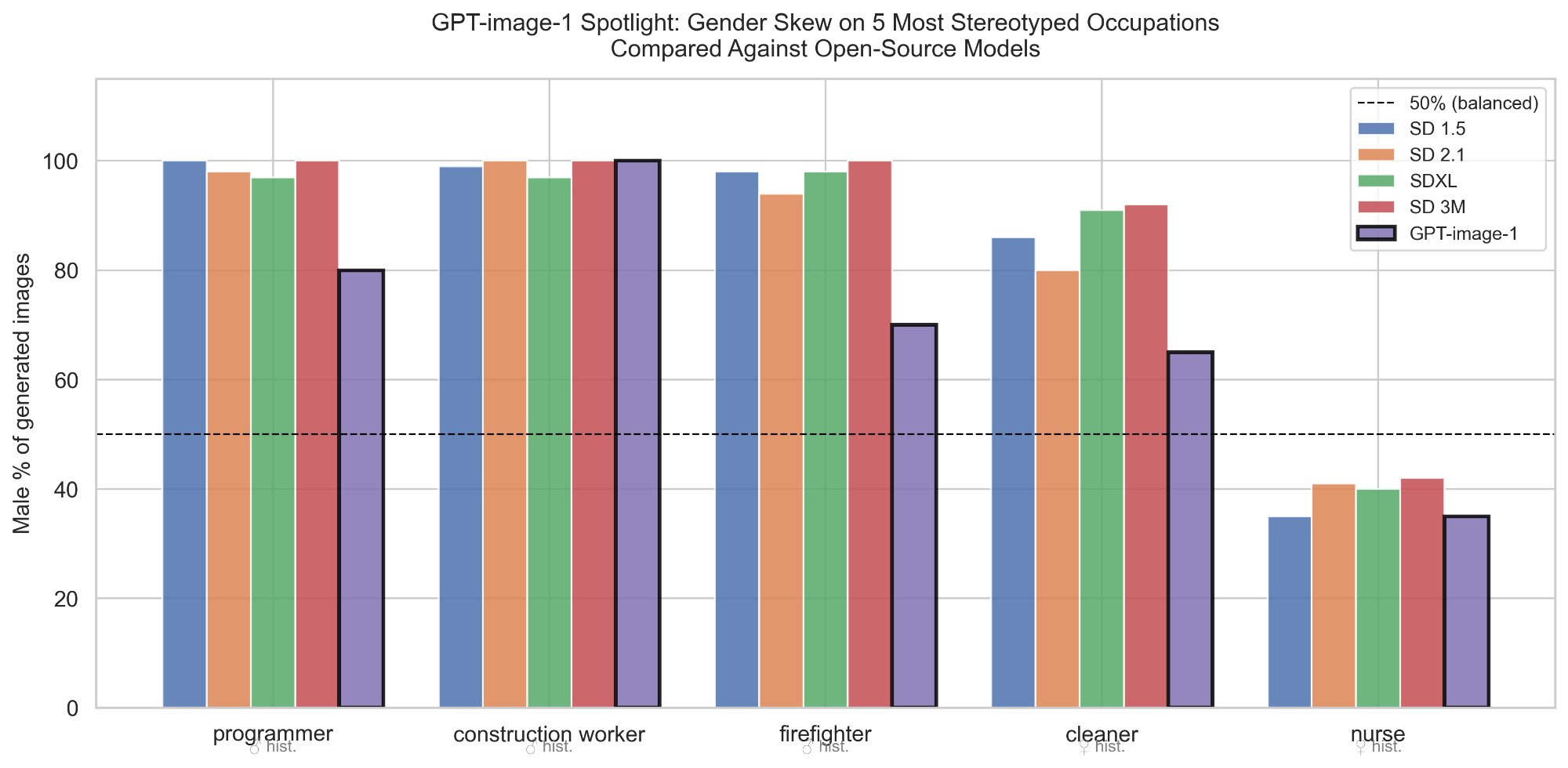}
\caption{Preliminary GPT-image-1 comparison (purple, bold border) vs open-source models on 5 occupations. Cram\'er's $V = 0.080$ (small effect). $n=20$ for GPT-image-1 per occupation; CIs are approximately $\pm$20pp.}
\label{fig:gpt}
\end{figure}

GPT-image-1 generates 80\% male images for programmer (vs.\ 97--100\% for open-source models), 70\% for firefighter (vs.\ 94--100\%), and 65\% for cleaner (vs.\ 80--92\%). For nurse, GPT-image-1 at 35\% male is comparable to SD 1.5 (35\%) and better than SD 2.1, SDXL, and SD 3 Medium (40--42\%). Construction worker is 100\% male across all models. Figure~\ref{fig:gpt} shows the comparison. Note that three GPT-image-1 cells---cleaner (65\% $\pm$20.9pp), firefighter (70\% $\pm$20.1pp), and nurse (35\% $\pm$20.9pp)---have wide CIs due to $n=20$, and should be treated as indicative. A proper follow-up would apply the same multi-prompt, multi-occupation design used for the open-source models.

\begin{table}[H]
\centering
\caption{Male percentage on 5 spotlight occupations. GPT-image-1 results are preliminary ($n=20$, single prompt, Cram\'er's $V = 0.080$).}
\label{tab:gpt}
\begin{tabular}{lccccc c}
\toprule
\textbf{Occupation} & \textbf{SD 1.5} & \textbf{SD 2.1} & \textbf{SDXL} & \textbf{SD 3M} & \textbf{GPT-1*} & \textbf{Hist.} \\
\midrule
Programmer & 100 & 98 & 97 & 100 & 80 & $\male$ \\
Construction Worker & 99 & 100 & 97 & 100 & 100 & $\male$ \\
Firefighter & 98 & 94 & 98 & 100 & 70$^\dagger$ & $\male$ \\
Cleaner & 86 & 80 & 91 & 92 & 65$^\dagger$ & $\female$ \\
Nurse & 35 & 41 & 40 & 42 & 35$^\dagger$ & $\female$ \\
\bottomrule
\end{tabular}
\\[4pt]
\raggedright\footnotesize $\male$ = historically male-skewed. $\female$ = historically female-skewed. Open-source: $n=100$ per occupation. *GPT-image-1: $n=20$ per occupation, single prompt. $^\dagger$ CI $\approx \pm$20pp; treat as indicative. Overall: $\chi^2 = 15.61$, BH-adjusted $p<0.001$, Cram\'er's $V=0.080$.
\end{table}

\section{Discussion}

\subsection{Large and Robust Deviations From Workforce Demographics}

The primary findings are robust regardless of where you look. At the model level, CIs are tight ($\pm$1.8pp) and the 76.4\% overall male rate sits 26pp above the 50\% baseline. At the occupation level, the most important results---programmer, construction worker, firefighter, engineer, and scientist---all have lower CI bounds well above 50\% despite the $\pm$9.8pp occupation-level CI. The amplification gaps for near-balanced occupations ($-30$ to $-47$pp for scientist, $-26$ to $-42$pp for cleaner) are three to five times the CI width and cannot be attributed to sampling variation.

Of 80 occupation-model cells, 16 fall within the $\pm$9.8pp CI of 50\% and are treated as indicative rather than definitive. The most notable are SD 3 Medium's teacher (45\% male) and social worker (45\% male), which straddle 50\% and indicate improvement relative to earlier models but cannot be called definitively female-majority without a larger sample.

\subsection{The Deployment Gap}

The non-linear trajectory---SD 1.5 and SD 2.1 statistically indistinguishable ($V=0.004$), SDXL significantly worse than both ($V=0.050$ and $0.045$ respectively), SD 3 Medium significantly better than SDXL ($V=0.126$)---reveals what we term a deployment gap: a situation in which a widely adopted model version shows higher bias than both its predecessor and its successor. Users relying on SDXL get no benefit from SD 3 Medium's improvements unless they actively switch, and there is no reason to assume they know to.

\subsection{Cross-Directional Bias Is Systematic}

57.6\% of images for historically female occupations are classified as male---a reversal that is statistically overwhelming (raw $p = 3.43 \times 10^{-22}$, BH-adjusted $p = 1.71 \times 10^{-21}$, $n=4{,}000$) is robust. SD 3 Medium shows meaningful improvement on this dimension ($V=0.199$ vs.\ SDXL on female-coded occupations) but the improvement is selective---teacher and social worker improve while pilot regresses---suggesting targeted rather than systematic correction.

\subsection{Prompt Sensitivity Grows With Generation}

More capable models are less consistent in their demographic representations across prompt phrasings, not more. Female-coded occupations show consistently higher sensitivity than male-coded ones, suggesting less stable learned representations for these roles. Evaluating a model on one prompt per occupation misses most of the variation it can produce---a practical reason to test multiple phrasings in any bias audit.

\subsection{GPT-image-1: A Preliminary Observation}

The GPT-image-1 comparison yields a small practical effect (Cram\'er's $V=0.080$) despite statistical significance (BH-adjusted $p < 0.001$). Three of five GPT-image-1 cells have CIs of approximately $\pm$20pp due to $n=20$, making those occupation-level comparisons indicative only. A proper follow-up would apply the same multi-prompt, multi-occupation design used for the open-source models.

\subsection{Practical Recommendations}

Based on these findings:
\begin{itemize}
    \item \textbf{Evaluate each model version independently.} Bias does not improve monotonically. SDXL is more biased than SD 1.5. Every deployed model version should be evaluated for demographic representation.
    \item \textbf{Test multiple prompt phrasings.} A single prompt underestimates output range. Evaluate at least three to five phrasings across all relevant occupations.
    \item \textbf{Compare against demographic ground truth.} Comparing outputs against BLS or equivalent national workforce data reveals deviations that internal 50/50 balance metrics do not capture.
\end{itemize}

\section{Limitations}

\textbf{Statistical power and confidence intervals.} Each occupation-model cell contains $n=100$ images (5 prompts $\times$ 20), giving 95\% CIs of approximately $\pm$9.8pp at $p=0.5$. Model-level estimates ($n=2{,}000$) have CIs of approximately $\pm$1.8pp. Of 80 open-source occupation-model cells, 16 fall within $\pm$9.8pp of 50\% and are flagged as indicative throughout. Multiple comparisons are corrected using BH across all 10 tests; all nine significant findings survive correction.

\textbf{Classifier validation.} DeepFace was trained on real photographs. Our single-reviewer manual validation on 50 images (96\% agreement) is limited in sample size and lacks inter-rater reliability. We argue the primary findings involving gaps of 26pp or more above baseline are robust to plausible classifier error, but occupation-level findings near 50\% should be interpreted with caution.

\textbf{Binary gender classification.} DeepFace classifies apparent gender as Man or Woman, missing non-binary, non-conforming, and androgynous presentations. Apparent gender is a proxy for perceived presentation, not self-identified gender.

\textbf{Resolution differences.} Open-source models were run at 512$\times$512 pixels while GPT-image-1 images were generated at 1024$\times$1024. DeepFace face detection and attribute classification accuracy may differ across resolutions. This provides a further reason to treat the GPT-image-1 comparison as exploratory, as some differences between GPT-image-1 and open-source outputs could reflect resolution-driven classification variation rather than model behavior alone.

\textbf{Default inference settings and ecological validity.} All models were run with default settings and no negative prompts. Many users of Stable Diffusion models employ negative prompts (e.g., ``bad anatomy, blurry, low quality'') and modified guidance scales that can substantially affect demographic outputs. Results reflect out-of-the-box behavior rather than the full distribution of real-world use, which limits ecological validity.

\textbf{BLS data specificity and occupational mapping.} Three occupational mappings involve meaningful ambiguity. ``Cleaner'' maps to a combined estimate across janitors and building cleaners (SOC 37-2011, $\sim$29\% female) and maids and housekeeping cleaners (SOC 37-2012, $\sim$89\% female), which have very different gender compositions; the 46\% figure used is a weighted average. ``Babysitter'' has no BLS SOC code and maps to childcare workers (SOC 39-9011, 94\% female). ``Scientist'' maps to the broad category of life, physical, and social science occupations (SOC 19-0000, 48\% female), which aggregates subfields ranging from $\sim$20\% female (physics) to $\sim$75\% female (psychology). These ambiguities affect the precision of the BLS reference values for these three occupations but do not change the direction of the findings. BLS figures reflect U.S.\ 2023 patterns and may not transfer to other cultural contexts.

\textbf{Occupational scope.} Twenty occupations is a reasonable but non-exhaustive starting point. Gender-neutral and emerging roles are not included.

\textbf{Model access.} SD 3.5 and later variants were not evaluated due to hardware constraints.

\textbf{GPT-image-1 comparison.} Five occupations, one prompt, $n=20$ per occupation; three cells have CIs $\approx \pm$20pp. Combined with the resolution difference from open-source models, Cram\'er's $V = 0.080$ should be treated as a preliminary observation only.

\section{Conclusion}

Across 8,000 generated images and four model generations, one pattern holds without exception: models consistently depict professional roles as more male than the actual workforce---and newer models are not reliably fairer. We measured this systematically across 20 occupations, 5 prompt phrasings, and four Stable Diffusion generations, with $n=100$ per occupation-model cell, confidence intervals throughout, and Benjamini--Hochberg correction across all 10 tests.

Male dominance is pervasive and statistically robust: 76.4\% of 8,000 open-source images show male subjects (95\% CI [75.1\%, 78.7\%], $p < 2.2 \times 10^{-16}$, BH-adjusted). 57.6\% of images for historically female occupations show male subjects (raw $p = 3.43 \times 10^{-22}$, BH-adjusted $p = 1.71 \times 10^{-21}$). Models underrepresent women by 20--46pp on average relative to BLS workforce demographics, with deviations of 26--47pp for near-balanced occupations such as scientist and cleaner.

Bias does not improve steadily across generations. SDXL---one of the most broadly adopted open-source Stable Diffusion variants---shows higher gender skew than both SD 1.5 and SD 3 Medium ($V=0.050$ and $0.126$ respectively). We introduce the term \emph{deployment gap} to describe this pattern. A preliminary comparison with GPT-image-1 suggests lower bias on four of five spotlight occupations ($V=0.080$, small effect), warranting follow-up study. Code and generation configurations are available at \url{https://github.com/SheshNGupta/GenderSterotype}.

\section*{Ethics Statement}

This study does not involve human subjects. All images analyzed were generated by AI models in response to occupational text prompts; no images of real individuals were collected, used, or stored. DeepFace classifications assign apparent gender labels (Man or Woman) to synthetic faces. We acknowledge that this binary classification does not capture the full spectrum of gender identity and presentation, and we discuss this limitation explicitly in Section~6.

The images generated in this study depict professional roles and were used solely to measure demographic patterns in model outputs. No images were generated with the intent to produce harmful, demeaning, or discriminatory content. Raw generated images are not released publicly to avoid potential misuse; only aggregate statistics and figures derived from them are shared.

The findings of this paper document gender bias in widely deployed AI systems. We believe transparency about these patterns is necessary for the research community and for developers to address them. We are aware that detailed documentation of model biases could in principle be used to exploit or exacerbate those biases, but we judge this risk to be low given that the patterns we identify are already observable to any user of these systems. The benefit of public documentation substantially outweighs this risk.

All generation configurations, classification scripts, and aggregate results are made publicly available to support reproducibility and future audits.

\section*{Acknowledgments}

The authors thank the developers of the Hugging Face Diffusers library and the DeepFace framework, which made this large-scale evaluation feasible on consumer hardware.

\bibliographystyle{unsrt}
\bibliography{references}

@inproceedings{bolukbasi2016man,
  title={Man is to computer programmer as woman is to homemaker? {D}ebiasing word embeddings},
  author={Bolukbasi, Tolga and Chang, Kai-Wei and Zou, James Y. and Saligrama, Venkatesh and Kalai, Adam T.},
  booktitle={Advances in Neural Information Processing Systems},
  volume={29},
  year={2016}
}

@inproceedings{zhao2017men,
  title={Men also like shopping: Reducing gender bias amplification using corpus-level constraints},
  author={Zhao, Jieyu and Wang, Tianlu and Yatskar, Mark and Ordonez, Vicente and Chang, Kai-Wei},
  booktitle={Proceedings of the 2017 Conference on Empirical Methods in Natural Language Processing (EMNLP)},
  pages={2979--2989},
  year={2017}
}

@inproceedings{bianchi2023easily,
  title={Easily accessible text-to-image generation amplifies demographic stereotypes at large scale},
  author={Bianchi, Federico and Kalluri, Pratyusha and Durmus, Esin and Ladhak, Faisal and Cheng, Myra and Nozza, Debora and Hashimoto, Tatsunori and Jurafsky, Dan and Zou, James and Caliskan, Aylin},
  booktitle={Proceedings of the 2023 ACM Conference on Fairness, Accountability, and Transparency (FAccT)},
  pages={1493--1504},
  year={2023}
}

@inproceedings{cho2023dalleval,
  title={{DALL-Eval}: Probing the reasoning skills and social biases of text-to-image generation models},
  author={Cho, Jaemin and Zala, Abhay and Bansal, Mohit},
  booktitle={Proceedings of the IEEE/CVF International Conference on Computer Vision (ICCV)},
  pages={3043--3054},
  year={2023}
}

@article{mandal2023multimodal,
  title={Multimodal composite association score: Measuring gender bias in generative multimodal models},
  author={Mandal, Abhishek and Leavy, Susan and Little, Suzanne},
  journal={arXiv preprint arXiv:2304.13855},
  year={2023}
}

@misc{blsdata2023,
  title={Labor Force Statistics from the Current Population Survey, Table 11: Employed persons by detailed occupation, sex, race, and Hispanic or Latino ethnicity},
  author={{U.S. Bureau of Labor Statistics}},
  year={2023},
  howpublished={\url{https://www.bls.gov/cps/cpsaat11.htm}},
  note={Accessed: March 2025}
}

@inproceedings{torralba2011unbiased,
  title={Unbiased look at dataset bias},
  author={Torralba, Antonio and Efros, Alexei A.},
  booktitle={Proceedings of the 2011 IEEE Conference on Computer Vision and Pattern Recognition (CVPR)},
  pages={1521--1528},
  year={2011},
  organization={IEEE}
}

@inproceedings{buolamwini2018gender,
  title={Gender shades: Intersectional accuracy disparities in commercial gender classification},
  author={Buolamwini, Joy and Gebru, Timnit},
  booktitle={Proceedings of the 1st Conference on Fairness, Accountability and Transparency},
  volume={81},
  pages={77--91},
  year={2018},
  organization={PMLR}
}

@inproceedings{oppenlaender2022creativity,
  title={The creativity of text-to-image generation},
  author={Oppenlaender, Jonas},
  booktitle={Proceedings of the 25th International Academic Mindtrek Conference},
  pages={192--202},
  year={2022}
}

@article{weidinger2021ethical,
  title={Ethical and social risks of harm from language models},
  author={Weidinger, Laura and Mellor, John and Rauh, Maribeth and Griffin, Conor and Uesato, Jonathan and Huang, Po-Sen and Cheng, Myra and Glaese, Mia and Balle, Borja and Kasirzadeh, Atoosa and Kenton, Zac and Brown, Sasha and Hawkins, Will and Stepleton, Tom and Biles, Courtney and Birhane, Abeba and Haas, Julia and Rimell, Laura and Hendricks, Lisa Anne and Isaac, William and Legassick, Sean and Irving, Geoffrey and Gabriel, Iason},
  journal={arXiv preprint arXiv:2112.04359},
  year={2021}
}

@inproceedings{serengil2021hyperextended,
  title={{HyperExtended LightFace}: A facial attribute analysis framework},
  author={Serengil, Sefik Ilkin and {\"O}zpinar, Alper},
  booktitle={2021 International Conference on Engineering and Emerging Technologies (ICEET)},
  pages={1--4},
  year={2021},
  organization={IEEE}
}

@article{luccioni2023stable,
  title={Stable bias: Analyzing societal representations in diffusion models},
  author={Luccioni, Alexandra Sasha and Akiki, Christopher and Mitchell, Margaret and Jernite, Yacine},
  journal={Advances in Neural Information Processing Systems},
  volume={36},
  year={2023}
}

@article{friedrich2023fair,
  title={Fair diffusion: Instructing text-to-image generation models on fairness},
  author={Friedrich, Felix and Schramowski, Patrick and Brack, Manuel and Struppek, Lukas and Hintersdorf, Dominik and Luccioni, Sasha and Kersting, Kristian},
  journal={arXiv preprint arXiv:2302.10893},
  year={2023}
}

\end{document}